\documentclass[runningheads]{llncs}

\usepackage[utf8]{inputenc} 
\usepackage[T1]{fontenc}    
\usepackage{hyperref}       
\usepackage{url}            
\usepackage{booktabs}       
\usepackage{amsfonts}       
\usepackage{nicefrac}       
\usepackage{microtype}      
\usepackage{xcolor}         
\usepackage{graphicx}
\usepackage{adjustbox}
\usepackage{amsmath}
\usepackage{algorithm}
\usepackage{algorithmicx}
\usepackage[noend]{algpseudocode}
\usepackage{soul}

\DeclareMathOperator*{\argmin}{arg\,min}

\title{Random Dilated Shapelet Transform: A New Approach for Time Series Shapelets}
\titlerunning{Random Dilated Shapelet Transform}

\author{Antoine Guillaume\inst{1,2}\orcidID{0000-0001-8975-982X} \and
Christel Vrain\inst{1}\orcidID{0000-0003-3307-0753} \and
Wael Elloumi\inst{2}\orcidID{0000-0002-5886-2450}}
\authorrunning{A. Guillaume et al.}
\institute{Univ. Orléans, INSA Centre Val de Loire, LIFO EA 4022, F-45067, Orléans, France \and
Worldline\\
\email{antoine.guillaume@worldline.com, christel.vrain@univ-orleans.fr, wael.elloumi@worldline.com}\\
}

\begin{document}

\maketitle

\begin{abstract}
Shapelet-based algorithms are widely used for time series classification because of their ease of interpretation, but they are currently outperformed by recent state-of-the-art approaches. We present a new formulation of time series shapelets including the notion of dilation, and we introduce a new shapelet feature to enhance their discriminative power for classification. Experiments performed on 112 datasets show that our method improves on the state-of-the-art shapelet algorithm, and achieves comparable accuracy to recent state-of-the-art approaches, without sacrificing neither scalability, nor interpretability.
\end{abstract}
\keywords{Time series \and Shapelets \and Classification}
\section{Introduction}
Time series occur in a multitude of domains, covering a wide range of applications, which have impacts in many parts of society. The ever-increasing quantity of data and the publications of laws regarding models interpretability are setting new constraints for applications across industries. 

Recent research in time series classification produced highly accurate classifiers, using either deep learning approaches \cite{Inception}, or meta-ensemble methods \cite{HC1,HC2}. Despite being the most accurate approaches, they are among the slowest, which make them hard to apply on use-cases with huge amount of data. 
Other methods, based on random approaches, notably the RandOm Convolutional KErnel Transform (ROCKET) \cite{Rocket}, achieve comparable accuracy with extreme scalability. Even though recent works on post-hoc methods and specific frameworks \cite{XAITS} improved the interpretability of those approaches, they lack a "by design" interpretability.

On the other hand, time series shapelets \cite{Shapelets} have been widely used in time series classification for their ease of interpretation, which is a critical aspect to some application domains such as health and security. 
The downside is that shapelet algorithms are often outperformed by recent approaches, both in terms of accuracy and scalability. Most Shapelet approaches tried to solve the scalability issues at the expense of some classification accuracy, notably through the use of symbolic approximation techniques \cite{FastShp}, while others used random shapelets \cite{UFShp}. Recently, a symbolic sequence ensemble learning \cite{MrSEQL} method was proposed, which improved the predictive power of approximation-based methods, while other work focused on finding a new discriminative feature \cite{LRS} to consider during the extraction process.

In this work, we present the Random Dilated Shapelet Transform, an adaptation of time series shapelets that includes the notion of dilation, one of the core mechanism of the success of convolutional kernel approaches. We also extend on the work of \cite{LRS} and introduce a new feature to enhance the discriminative power of shapelets. Our contributions can be summarized as follows:

\begin{itemize}
\item an adaptation of time series shapelets allowing the use of dilation, and a feature to capture a new discriminative property of shapelets, 
\item an interpretable, scalable and accurate shapelet algorithm, which allows shapelet based algorithm to catch-up with the state-of-the-art,
\item an experimental study about the sensitivity of our method parameters and a comparative study against the state-of-the-art algorithms for time series classification.
\end{itemize}

\section{Background}

In this section, we present a brief review of time series classification, and make a focus on shapelet methods. From now on, we use calligraphic font (\({\cal X}\)) to denote a collection of elements (e.g. a set of time series), capital (\(X\)) for one element (e.g. a time series), and lowercase (\(x\)) for a value of this element. In this work we consider supervised classification: the ensemble of input time series will be denoted by \({\cal X}=\{X_1, ..., X_n\}\) with \(X_i=\{x_1, ..., x_m\}\) a time series and \(Y=\{y_1, ..., y_n\}\) their respective classes.

\subsection{Time series classification}
\label{sec:TSC}
We present a brief overview of the algorithms identified as state-of-the-art and used in our experimental section, and we report the reader to a recent review \cite{TSCreview} for a more detailed view of the field.
\begin{itemize}
\item \textbf{Shapelet Transform Classifier (STC)} \cite{STC}, is regarded as a state of the art for shapelet algorithms in terms of accuracy. This algorithm iteratively initializes new shapelets, assesses their discriminative power, and removes those that are too similar. The goal being to maximize the discriminative power of an ensemble of shapelets. A Rotation Forest is then applied as a classifier.
\item \textbf{Temporal Dictionary Ensemble (TDE)} \cite{TDE} is an ensemble of dictionary-based classifiers. It uses some variants of the BOSS classifier \cite{BOSS} and WEASEL \cite{WEASEL}, as base estimators and optimizes their parameters through a Gaussian process.
\item \textbf{Diverse Representation Canonical Interval Forest Classifier (DrCIF)} \cite{HC2}, is an extension of the CIF algorithm \cite{CIF}. After selecting random intervals from different representations of the time series, it uses the Catch22 \cite{catch22} method  to extract a feature matrix.
\item \textbf{RandOm Convolutional KErnel Transform (ROCKET)} \cite{Rocket}, randomly generates a huge set of convolutional kernels, and extracts as features the maximum and the proportion of positive values of the convolution for each time series and each kernel. It is followed by a Ridge classifier. An ensemble version of this method, called ARSENAL, was introduced by \cite{HC2}.
\item \textbf{Inception-Time} \cite{Inception} is an ensemble of Inception networks, which introduce Inception modules as replacement for traditional fully convolutional layers, notably to mitigate the vanishing gradient problem.
\item \textbf{Hierarchical Vote Collective of Transformation-based Ensembles (HC1)} \cite{HC1}is a meta-ensemble classifier using a variety of time series classifiers, such as STC, with a novel ensemble learning scheme which estimate the weight of each base classifier in the final decision. This estimation is based on performance in a 10-fold validation scheme.

Variants of this method were developed, such as TS-CHIEF \cite{TSCHIEF} and HC2 \cite{HC2}, that both modify the set of base classifiers to improve accuracy. HC2 also modified the meta-ensemble procedure, using a Out-Of-Bag estimate instead of a 10-fold validation to estimate the performance of each base classifier, which improved scalability compared to HC1.
\end{itemize}

\subsection{Shapelets}
\label{sec:shapelet}
Shapelets \cite{Shapelets} were originally defined as time series subsequences representative of class membership. In the following, we define a shapelet $S$ as a vector $S=\{s_1, ..., s_l\}$ with $l$ its length. All shapelet-based algorithms have the same protocol to extract features from a shapelet $S$ and a time series $X=\{x_1, ..., x_m\}$, by using a distance vector $f(S,X) = \{f_1, ..., f_{m-(l-1)}\} $ defined as :

\begin{equation}
f_i = \sqrt{\sum_{j=1}^{l} (X_{i+(j-1)} - s_j)^2}
\end{equation}

In this definition, a point $f_i$ is simply the Euclidean distance between $S$ and the subsequence of length $l$ starting at index $i$ in $X$. The minimum value of $f(S,X)$ is then extracted as a feature, which can be interpreted as an indicator of the presence of the pattern represented by $S$ in $X$. A popular variant of this distance function consists in using a z-normalized Euclidean distance, where $S$ and all subsequences of $X$ are z-normalized independently, allowing to add scale invariance to the translation invariance of the initial formulation.
Then, as presented in the Shapelet Transform \cite{ST}, by using a set of shapelets $\cal S$, one can then transform an ensemble of time series $\cal X$ into a feature matrix of shape $(|{\cal X}|,|{\cal S}|)$, and use it as input in a non-temporal classifier such as a decision tree.\newline

The step of generating and selecting shapelet candidates is the main difference between most approaches. In order to speed up the exhaustive search, Fast Shapelet \cite{FastShp} use input discretization, while Ultra Fast Shapelet \cite{UFShp} use random sampling. FLAG \cite{FLAG} build shapelets location indicators from the data to reduce the set of admissible candidates, and GENDIS \cite{GENDIS} use an evolutionary algorithm initialized by a clustering on the set of possible candidates. Learning Time Series Shapelet \cite{LTS} use a gradient-descent optimization that iteratively change the values of a set of shapelets. MrSEQL \cite{MrSEQL}, while not strictly speaking a shapelet algorithm, searches for discriminative symbolic sequences in a variety of symbolic representations of the inputs.

Since the publication of Localized Random Shapelet (LRS) \cite{LRS}, which showed the benefit of extracting $argmin\ d(S,X)$ to discriminate time series based on the location of the minimum between $S$ and $X$, it has been included in most recent approaches. Based on their results, we will also use this feature in our method.
\section{Proposed method}
\label{sec:RDST}
In this section, we introduce the main components of our method: the use of dilation in the shapelet formulation and the features extracted from the distance vector between a shapelet and a time series. We put emphasis on the dilation and on the Shapelet Occurrence feature that are new contributions to shapelet algorithms. We give some simple visual examples to illustrate these notions, and report the visualization on real data to the experimental section.

\subsection{Dilated Shapelets}
To introduce the notion of dilation in shapelets, we define now a shapelet $S$ as $S=\{ \{v_1, ..., v_l\}, d \}$ with $l$ the length parameter and $d$ the dilation parameter. In practice, the dilation is used in the distance function $f$, where each value of the shapelet will be compared to a dilated subsequence of the input time series. More formally, consider a time series $X=\{x_1, ..., x_m\}$ and a dilated shapelet $S$, we now define $f(S,X) = \{f_1, ..., f_{m-(l-1)\times d}\}$ as :
\begin{equation}
f_i = \sqrt{\sum_{j=1}^{l} (X_{i+(j-1) \times d} - s_j)^2}
\end{equation}

\begin{figure}[h]
  \includegraphics[width=0.9\textwidth]{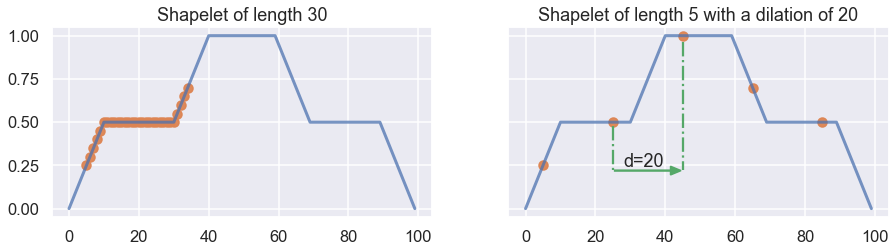}
  \centering
  \caption{An example of two possible shapelets (in orange), positioned on a synthetic pattern (in blue): (a) one without dilation, and (b) a much smaller one but with dilation}
  \label{fig:ex_dil_shp}
\end{figure}

The interest of using dilation in shapelets is to make them non-contiguous subsequences. It allows a shapelet to either match a non-contiguous pattern, or a contiguous one, by focusing on key points of the pattern without covering it entirely, as illustrated in Figure \ref{fig:ex_dil_shp}. Note that this formulation is equivalent to the original shapelet formulation when $d=1$.

\subsection{Shapelet Occurrence feature}
\label{sec:SO}
If we consider a shapelet $S$ and two time series $X_1$ and $X_2$, we can imagine multiple ways of discriminating $X_1$ and $X_2$ using $S$.
\begin{itemize}
\item $S$ can be present in $X_1$ but not in $X_2$. This is captured by $min\ f(S,X_i)$, with smaller distances indicating better matches between the shapelet and the series.
\item $S$ can be present in both series, but not at the same place. This is captured by the $argmin$ feature introduced by LRS \cite{LRS}.
\item $S$ can be present in both series, but not at the same scale. In this case, a normalized distance would not be able to discriminate the series.
\item $S$ can be present in both series, but occurs a different number of times in $X_1$ compared to $X_2$. This is captured by a new feature, called Shapelet Occurrence (SO).
\end{itemize}

\begin{figure}[h]
  \includegraphics[width=1.\textwidth]{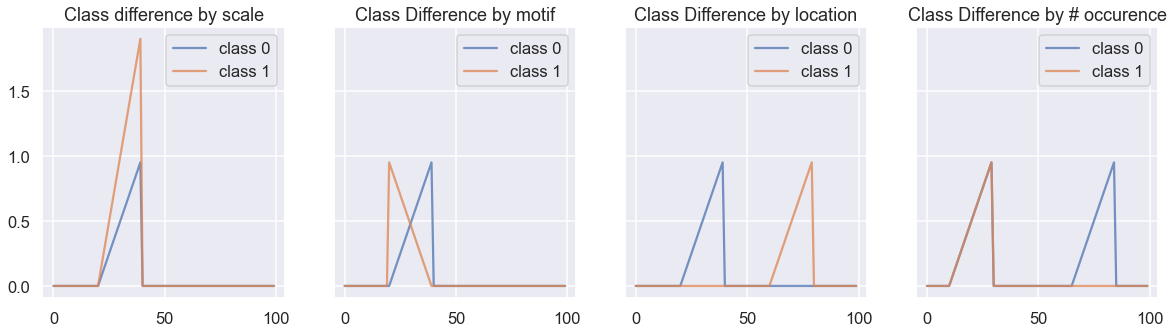}
  \centering
  \caption{Synthetic examples of possible discriminative properties between two classes}
  \label{fig:shp_disc}
\end{figure}

Those points are illustrated in Figure \ref{fig:shp_disc}. Deciding whether scaling is important or not is highly dependent on the application, but without prior knowledge, one cannot know which to choose. For this reason, we introduce a parameter in Section \ref{sec:RandDST} allowing to tune the amount of normalized shapelets.

To the best of our knowledge, the number of occurrences of a shapelet has never been considered as a feature. This requires another modification to the definition of $S$ as $S=\{ \{v_1, ..., v_l\}, d, \lambda \}$, with $\lambda$ a threshold allowing us to compute the Shapelet Occurrence ($SO$) feature as $SO = |\{i| f(S,X)_i <\lambda\}|$. Although the parameter $\lambda$ could be set randomly, we discuss in Section \ref{sec:RandDST} a method to set the value of this threshold.

\subsection{Random Dilated Shapelet Transform (RDST)}
\label{sec:RandDST}
Our objective for this algorithm is to produce an accurate but scalable approach. As our shapelet formulation adds attributes compared to the initial formulation \cite{Shapelets}, optimizing a set of dilated shapelets with a threshold $\lambda$ will be costly, and this explains why we choose a random approach.

For simplicity, we present our approach in the context of univariate and even length time series, with ${\cal X} = \{X_1, ..., X_n\}$  a set of time series ($X_i = \{x_1, ..., x_m \}$) and $Y=\{y_1, ..., y_n\}$ their respective classes. Our method takes as input four parameters that are: $n\_shapelets$ the number of shapelets to generate, $L$ a set of possible lengths for the shapelets, $p\_norm$ the proportion of shapelets that will use z-normalization, and $(P_1, P_2) \in [0,100]$ a pair used as percentile bounds for the sampling of the threshold $\lambda$. 

Given the definition of a shapelet $S=\{ \{v_1, ..., v_l\}, d, \lambda \}$, we initialize each parameter as follows:
\begin{itemize}
\item the length $l$ is uniformly drawn from $L$,
\item the dilation $d$, in the same way as ROCKET \cite{Rocket}, is set to $d = \left\lfloor 2^x \right\rfloor$ with $x$ uniformly drawn in $[0, log_2 \frac{m}{l}]$,
\item we randomly choose whether the shapelet will use a z-normalized distance with probability $p\_norm$,
\item for setting the values, a sample $X$ is uniformly drawn from ${\cal X}$, and an admissible start point $i$ (given $l, d$) is randomly selected. Then, values are set to $[X_i, ..., X_{i+(l-1)\times d}]$.
\item finally, given a shapelet $S$, to fix the value of $\lambda$,  we take a sample $X$ from the same class as the one used for extracting the shapelet value, and uniformly draw a value between the two percentiles $(P_1, P_2)$ of $f(S,X)$.
\end{itemize}

The strategy employed to find $\lambda$ is a classic trade-off between time and accuracy. If scalability was not a focus, we could compute the distance vector for more samples in ${\cal X}$, and optimize the value of $\lambda$ based on an information measure.
After computing the distance vector between all pairs of time series and shapelets, the output of our method is a feature matrix of size $(|{\cal X}|,3 \times n\_shapelets)$, with the three features extracted from the distance vector $f(S,X)$ being the $min, argmin, SO(S,X)$.

Following the arguments of the authors of ROCKET \cite{Rocket}, we use a Ridge Classifier after the transformation of ${\cal X}$, as the L2 regularization used in Ridge is of critical importance due to the high number of features that are generated, while being scalable and interpretable.
\section{Experiments}
\label{sec:Experiments}
Our focus in this section is to study the influence of the four parameters of our method on classification accuracy, as well as comparing its performance to recent state-of-the-art approaches. All the experiments were run on a DELL PowerEdge R730 on Debian 9 with 2 XEON E5-2630 Corei7 (92 cores) and 64GB of RAM. We provide a python package\footnote{https://github.com/baraline/convst} using community standards to run the method and the interpretability tool on any dataset, along with all result tables, and reproducibility instructions for our experiments.

In the following, we use the 112 univariate datasets from the UCR archive \cite{Dataset} and when comparing to state-of-the-art results, we use the same resamples scheme as the one used in their experiments. We use critical difference diagrams to display the mean ranks of objects, with cliques (formed by horizontal bars) computed using the Wilcoxon-Holm post-hoc analysis \cite{CriticalDiagram}, with a p-value of $0.05$. A clique indicates that the accuracy difference between objects is not statistically significant. 

\subsection{Sensitivity Analysis}
\begin{figure}[h]
  \includegraphics[width=1.\textwidth]{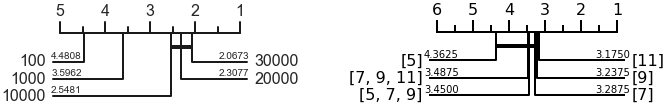}
  \centering
  \caption{Accuracy ranks for (a) different number of shapelets, and (b) different shapelet lengths}
  \label{fig:sensi_len_nshp}
\end{figure}

\begin{figure}[h]
  \includegraphics[width=1.\textwidth]{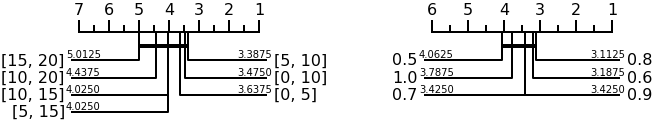}
  \centering
  \caption{Accuracy ranks for (a) different percentiles bounds, and (b) proportion of z-normalized shapelets}
  \label{fig:sensi_bound_pnorm}
\end{figure}

We conduct a sensitivity analysis on the four input parameters of our algorithm and their effect on classification accuracy on 40 datasets selected randomly, with raw results and selected datasets in a specific file in the online repository. 
For each parameter analysis, all other parameters remain fixed at the following default values : $n\_shapelets = 10000$, $p\_norm=0.9$, $L=[7,9,11]$, $P1=5$ and $P2=15$. Figure \ref{fig:sensi_len_nshp} and Figure \ref{fig:sensi_bound_pnorm} give the mean accuracy ranks of each method over the 40 datasets, with the accuracy of each method and each dataset computed as the mean of the same 10 resamples.
Given the tested set of values, the most impactful parameter is the number of shapelets, with a noticeable increase in performance above 10000 shapelets. All other parameters only display minor gains and thus seem to be stable. Based on those results, for all further experiments we set as default parameters $n\_shapelets = 10000$, $p\_norm=0.8$, $L=[11]$ and $P1=5$, $ P2=10$, and report results for datasets used in sensitivity analysis and the others.

\subsection{Scalability}
We perform a comparison of the scalability of our approach against Hive-Cote 1.0 (HC1), Hive-Cote 2.0 (HC2), DrCIF, ROCKET, and the Shapelet Transform Classifier (STC). Note that when used as a component in HC1 and HC2, STC is by default subject to a time contract of two hours. Except from this default configuration in HC1 and HC2, we are not setting any time contract in other algorithms. Both STC and RDST are by default sampling 10000 shapelets, and ROCKET use 10000 kernels.

We are aware that the runtime of HC1, HC2 and STC could be reduced with time contracts. But, as our goal in this section is to contextualize the gain in classification accuracy against the time complexity of each method, we present the results with the time contracts used to generate the accuracy results of the next section.

\begin{figure}[h]
  \includegraphics[width=1.\textwidth]{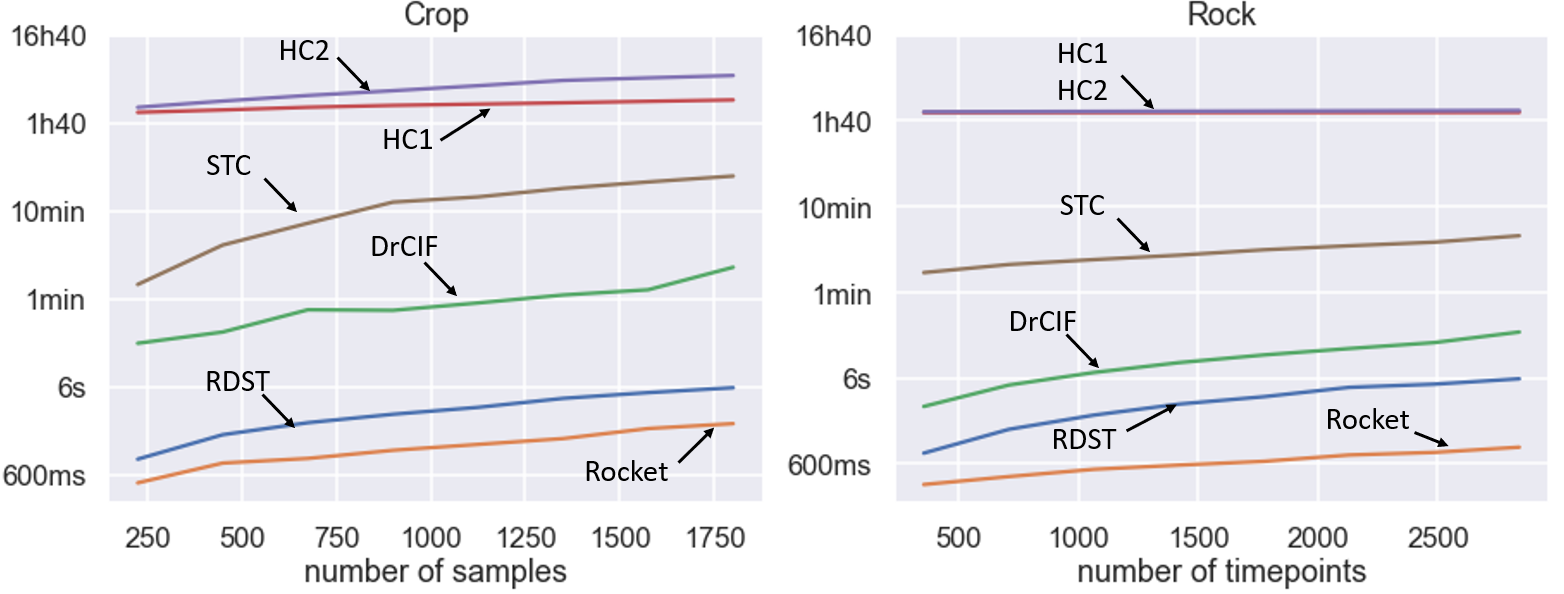}
  \centering
  \caption{Result of the scalability study of the competing algorithms for current state-of-the-art, for (a) number of time series and (b) time series length. Y-axis use log-scale.}
  \label{fig:scal}
\end{figure}

We use the Crop Dataset and the Rock Dataset of the UCR archive for evaluating the scalability respectively on the number of time series and their length. As all competing algorithms implemented in the sktime package of \cite{sktime} can use parallel processing, we set each algorithm to use 90 cores. Figure \ref{fig:scal} reports the mean training time over 10 resamples, showing the very competitive scalability of RDST. Note that due to job time limitation on our machine and the computational cost of HC2, we could not consider all samples for the Crop dataset. We report the reader interested in the implementation details of our algorithm to the web page of the project.

\subsection{Comparative study}
\begin{figure}[h]
  \includegraphics[width=1.0\textwidth]{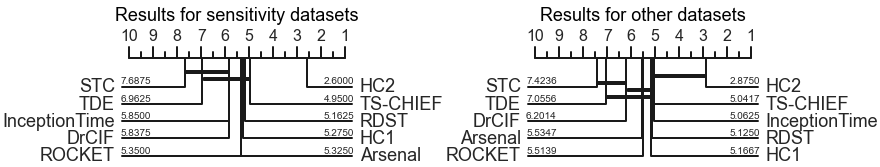}
  \centering
  \caption{Mean accuracy ranks of each method for the 40 dataset used in sensitivity analysis and the 72 others.}
  \label{fig:ranksdiv}
\end{figure}

We present the results of our comparative study using the mean accuracy over the same 30 resamples for each of the 112 datasets as HC2 \cite{HC2} used in their study, and compare our approach against their experimental result. Figure \ref{fig:ranksdiv} gives the mean accuracy rank of each method over the 40 datasets used for setting the defaults parameters in sensitivity analysis, and for the 72 others. The full result tables including standard deviation per dataset and more visualizations of the results are available online as supplementary materials. 

Given the scalability and simplicity of our method, having an accuracy comparable to the prior developments of HC2 and to deep learning approaches is a very promising result. Notably for future developments where focus would shift to accuracy rather than scalability. For reference, using RDST without any distance normalization is equivalent to STC in terms of mean accuracy rank, with the same protocol as above.

\subsection{Interpretability}
\begin{figure}[h!]
  \includegraphics[width=1.0\textwidth]{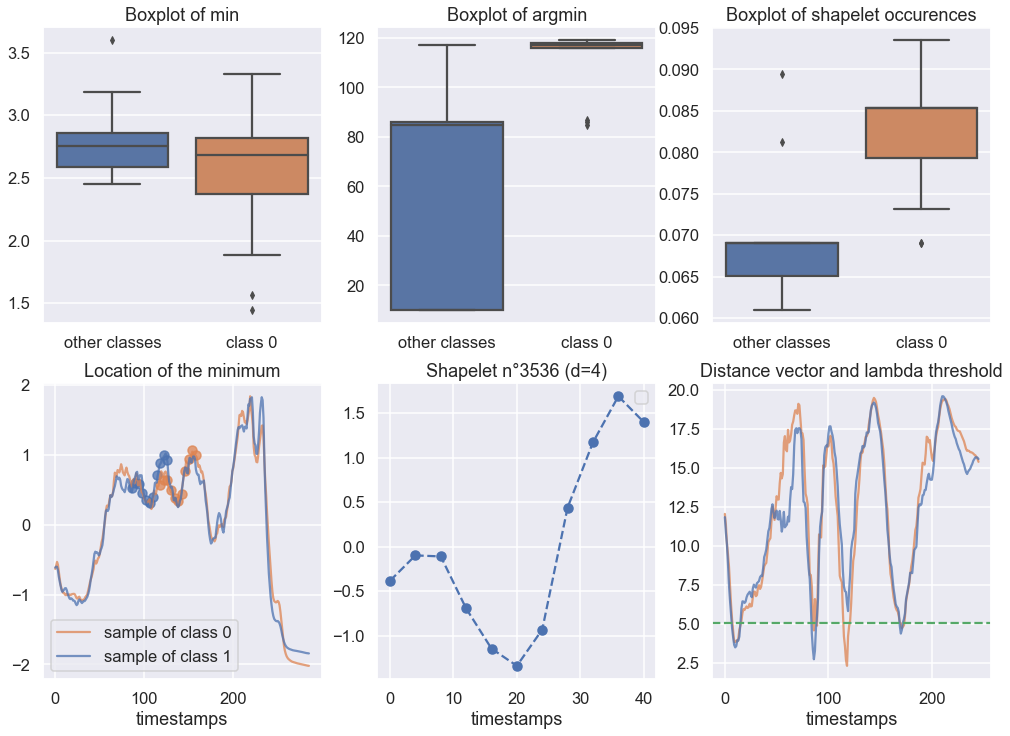}
  \centering
  \caption{The most important shapelet for class 0 of the Coffee dataset, according to weights of the Ridge classifier, with distribution displayed on the testing data, and two testing samples for visualization.}
  \label{fig:interp}
\end{figure}

\begin{figure}[h!]
  \includegraphics[width=1.\textwidth]{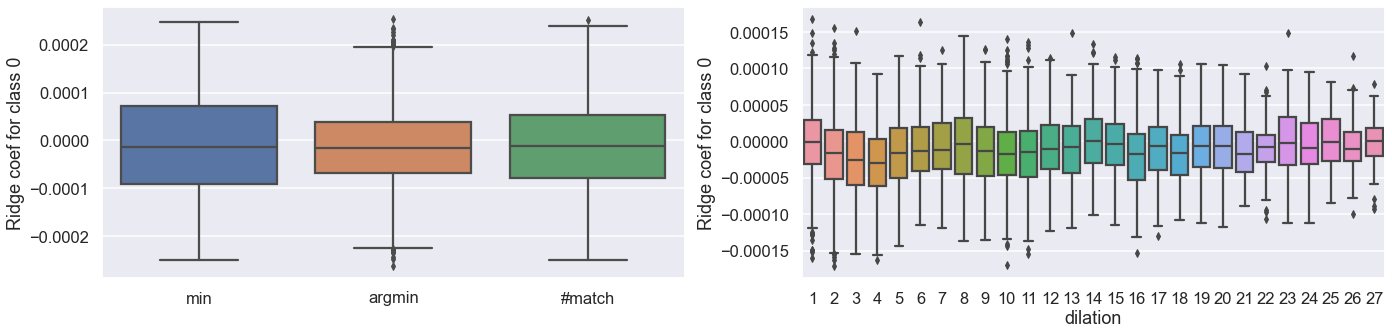}
  \centering
  \caption{A global interpretation of RDST, with (a) distribution of weights for each type of feature, and (b) distribution of weights per dilation.}
  \label{fig:interp2}
\end{figure}

Given a set of $M$ shapelets, RDST generates $3M$ features. Each feature is linked to a weight for each class in the Ridge classifier, as it is trained in a one-vs-all fashion. 
Given a class, we can then visualize either global or local information. Locally, we can inspect a shapelet to show how it discriminates the current class, and where the shapelet is positioned with either training or testing data, as shown in Figure \ref{fig:interp}. Globally, we can display the distribution of weights for each feature type ($min$, $\argmin$ and $SO$) or by shapelet characteristics such as length, dilation, or use of normalization as shown in Figure \ref{fig:interp2}. 
While this only provides a basic interpretation of the results, we believe a more formal framework could be developed to extract explanations from this data.

\section{Conclusions and future work}
The Random Dilated Shapelet Transform introduces new ways of increasing the global performance of shapelet algorithms, notably through the use of dilation, allowing the use of small non-contiguous subsequences as shapelets, efficiently covering areas of interest in the data. We have shown in our experiments that this new method improves on the state-of-the-art for shapelet algorithms with a good scalability compared to most of the approaches. This work offers many perspectives for future work, notably a generalized version to process uneven length or multivariate time series, as well as modifications of the shapelet generation process to better leverage class information. A more formal explainability framework is also one of our main priorities with this work, since being able to extract clear and visual explanations for domain experts is an extremely desirable property.
\subsubsection*{Acknowledgements}
This work is supported by the ANRT CIFRE grant n°2019/0281 in partnership with Worldline and the University of Orléans.

\bibliography{biblio}

\end{document}